\documentclass[10pt,twocolumn,letterpaper]{article}

\usepackage{cvpr}
\usepackage{times}
\usepackage{epsfig}
\usepackage{graphicx}
\usepackage{amsmath}
\usepackage{amssymb}

\usepackage{algorithm}
\usepackage[noend]{algpseudocode}

\usepackage[pagebackref=true,breaklinks=true,letterpaper=true,colorlinks,bookmarks=false]{hyperref}

\cvprfinalcopy 

\def\cvprPaperID{5608} 

\ifcvprfinal\fi
\begin{document}

\title{FilterReg: Robust and Efficient Probabilistic Point-Set Registration \\ using Gaussian Filter and Twist Parameterization}

\author{Wei Gao\\
Massachusetts Institute of Technology\\
{\tt\small weigao@mit.edu}
\and
Russ Tedrake\\
Massachusetts Institute of Technology\\
{\tt\small russt@mit.edu}
}

\maketitle

\begin{abstract}
Probabilistic point-set registration methods have been gaining more attention for their robustness to noise, outliers and occlusions. However, these methods tend to be much slower than the popular iterative closest point (ICP) algorithms, which severely limits their usability.
In this paper, we contribute a novel probabilistic registration method that achieves state-of-the-art robustness as well as substantially faster computational performance than modern ICP implementations. This is achieved using a rigorous yet computationally-efficient probabilistic formulation. Point-set registration is cast as a maximum likelihood estimation and solved using the EM algorithm. We show that with a simple augmentation, the E step can be formulated as a filtering problem, allowing us to leverage advances in efficient Gaussian filtering methods. We also propose a customized permutohedral filter~\cite{adams2010permu} for improved efficiency while retaining sufficient accuracy for our task. 
Additionally, we present a simple and efficient twist parameterization that generalizes our method to the registration of articulated and deformable objects. 
For articulated objects, the complexity of our method is almost independent of the Degrees Of Freedom (DOFs). 
The results demonstrate the proposed method consistently outperforms many competitive baselines on a variety of registration tasks. The video demo and source code are available on our  \href{https://sites.google.com/view/filterreg/home}{project page}.
\end{abstract}

\section{Introduction}
\label{sec:intro}

Point-set registration is the task of aligning two point clouds by estimating their relative transformation. This problem is an essential component for many practical vision systems, such as SLAM~\cite{nuchter20076dslam}, object pose estimation~\cite{lu1997poseestimation}, dense 3d reconstruction~\cite{whelan2016elasticfusion}, and interactive tracking of articulated~\cite{articulatedicp2007} and deformable~\cite{gao2018surfelwarp} objects. 

The ICP~\cite{icp1992} algorithm is the most widely used method for this task. ICP alternatively establishes nearest-neighbor correspondences and minimizes the point-pair distances. With spatial indices such as the KD-tree, ICP provides relatively fast performance. 
The literature contains many variants of the ICP algorithm; \cite{pomerleau2015review} and \cite{Pomerleau12comp} provide a thorough review and comparison.


Despite its popularity, the ICP algorithm is susceptible to noise, outliers and occlusions. These limitations have been widely documented in the literature~\cite{treeguass2018, cpd10, condem2011}. Thus, a great deal of research has been done on the use of probabilistic models for point-set registration~\cite{cpd10, granger2002multi, gold1998new}, which can in principle provide better outlier-rejection. 
Additionally, if each point is given a Gaussian variance, the point cloud can be interpreted as a Gaussian Mixture Model (GMM). Most statistical registration methods are built on the GMM and empirically provide improved robustness~\cite{cpd10, condem2011, evangelidis2014generative}. However, these methods tend to be much slower than the ICP and can hardly scale to large point clouds, which severely limits their practical usability. 

In this paper, we present a novel probabilistic registration algorithm that achieves state-of-the-art robustness as well as substantially faster computational performance than modern ICP implementations. 
To achieve it, we propose a computationally-efficient probabilistic model and cast the registration as a maximum likelihood estimation, which can be solved using the EM algorithm. With a simple augmentation, we formulate the E step as a filtering problem and solve it using advances in efficient Gaussian filters~\cite{adams2010permu, chen2007bilateralgrid, adams2009gaussian}. We also present a customized permutohedral filter~\cite{adams2010permu} with improved efficiency while retaining sufficient accuracy for our task. 
Empirically our method is as robust as state-of-the-art GMM-based methods, such as~\cite{cpd10}. In terms of the speed, our method with CPU is 3-7 times faster than modern ICP implementations and orders of magnitude faster than typical robust GMM-based methods. Furthermore, the proposed method can be GPU-parallelized and is 7 times faster than the CPU implementation. 

Additionally, we propose a simple and efficient twist parameterization that extends our method to articulated and node-graph~\cite{kavan2006dual} deformable objects. Our method is easy to implement and achieves substantial speedup over direct parameterization. For articulated objects, the complexity of our method is almost independent of the DOFs,  which makes it highly efficient even for high-DOF systems. Combining these components, we present a robust, efficient and general registration method that outperforms many competitive baselines on a variety of registration tasks. The video demo, supplemental document and source code are available on our  \href{https://sites.google.com/view/filterreg/home}{project page}.



\section{Related Work}

The problem of point set registration is extensively pursued in computer vision and an exhaustive review is prohibitive. In the following text, we limit our discussion to GMM-based probabilistic registration and review them roughly according to their underlying probabilistic models.

The earliest statistical methods~\cite{rangarajan1997robust, luo2003unified, mcneill2006probabilistic, granger2002multi} implicitly assumed the model points, which is controlled by the motion parameters (such as the rigid transformation or joint angles), induce a GMM distribution over the 3d space. The observation points are independently sampled from this distribution. Later, several contributions~\cite{cpd10, schulman2013tracking, condem2011} derived the EM procedure rigorously from the aforementioned probabilistic model. This formulation has also been applied to the registration of  multi-rigid~\cite{evangelidis2014generative}, articulated~\cite{ye2014real, condem2011} and deformable~\cite{cpd10, schulman2013tracking} objects. 

Another type of algorithms is known as the correlation-based methods~\cite{tsin2004correlation, jian2005robust, campbell2015adaptive, stoyanov2012ndtd2d}. These algorithms treat both observation points and model points as probabilistic distributions. The point-cloud registration can be interpreted as minimizing some distance between distributions, for instance the KL-divergence. To improve the efficiency, techniques such as voxelization~\cite{stoyanov2012ndtd2d} or Support Vector Machine~\cite{campbell2015adaptive} are used to create compact GMM representations.

In this paper, we assume that the observation points induce a probabilistic distribution over the space. Intuitively, the registration is to move the model points to positions with large posterior probability, subject to kinematic constraints. This formulation is related to several existing works~\cite{treeguass2018, magnusson2009three, cpd10}, and a more technical comparison is presented in Sec.~\ref{subsec:comparison}. In addition to the formulation, the key contribution of our work includes the introduction of the filter-based correspondence and twist parameterization built on the probabilistic model, as mentioned in Sec.~\ref{sec:intro}. Combining these components, the proposed method is general, robust and efficient that outperform various competitive baselines.

\vspace{-0.5em}
\section{Probabilistic Model for Registration}
\vspace{-0.3em}
\subsection{Probabilistic Formulation}
\label{subsec:probformulation}
\vspace{-0.2em}

In this subsection, we present our probabilistic model for point-set registration. We use $X, Y$ to denote the two point sets, $x_1, x_2, ..x_M$ and $y_1, y_2, ..., y_N$ are points in $X$ and $Y$. We define the \textbf{model} $X$ as the point set that is controlled by the \textbf{motion parameter} $\theta$. Another point set $Y$ is defined as the \textbf{observation}, which is fixed during the registration. 

\setlength{\abovedisplayskip}{3pt}
\setlength{\belowdisplayskip}{3pt}

We are interested in the joint distribution $p(X, Y, \theta)$. We assume given model geometry $X$, the observation $Y$ is independent of $\theta$, and the joint distribution $p(X, Y, \theta)$ can be factored as
\begin{equation}
   p(X, Y, \theta) \propto \phi_{\text{kinematic}}(X, \theta)\phi_{\text{geometric}}(X, Y)
\end{equation}

\noindent where $\phi_{\text{geometric}}(X, Y)$ is the potential function that encodes the geometric relationship, and the potential $\phi_{\text{kinematic}}(X, \theta)$ encodes the kinematic model. The $\phi_{\text{kinematic}}(X, \theta)$ can encode hard constraints such as $X = X(\theta)$ and/or soft motion regularizers, for instance the smooth terms in~\cite{cpd10, newcombe2015dynamicfusion} and the non-penetration term in ~\cite{schulman2013tracking}.

We further assume the kinematic model $\phi_{\text{kinematic}}(X, \theta)$ has already captured the dependency within model points $X$. Thus, \textit{conditioned on} the motion parameter $\theta$, the points in $X$ are independent of each other. The distribution can be further factored as 
\begin{equation}
   p(X, Y, \theta) \propto \phi_{\text{kinematic}}(X, \theta) \prod_{i=1}^{M} \phi_{\text{geometric}}(x_i, Y)
\end{equation}
A factor graph representation of our model is shown in Fig.~\ref{fig:formulation}. With these factorization schemes, the conditional distribution can be written as
\begin{equation}
   p(X, \theta | Y) \propto \phi_{\text{kinematic}}(X, \theta)\prod_{i=1}^{M} \phi_{\text{geometric}}(x_i | Y)
\end{equation}
Following several existing work~\cite{treeguass2018, cpd10}, we let the geometric distribution of each model point $\phi_{\text{geometric}}(x_i | Y)$ be a GMM,
\begin{equation}
\phi_{\text{geometric}}(x_i | Y) = \sum_{j=1}^{N+1} P(y_j)p(x_i|y_j)
\end{equation}
\noindent where $p(x_i|y_j) = \mathcal{N}(x_i;y_j, \Sigma_{xyz})$ is the Probability Density Function (PDF) of the Gaussian distribution, $y_j$ is the Gaussian centroid and $\Sigma_{xyz} = \text{diag}(\sigma_x^2, \sigma_y^2, \sigma_z^2)$ is the diagonal covariance matrix. An additional uniform distribution $p(x_i|y_{N+1}) = \frac{1}{M}$ is added to account for the noise and outliers. Similar to~\cite{cpd10}, we use equal membership probabilities $P(y_j) = \frac{1}{N}$ for all GMM components, and introduce a parameter $0 \leq w \leq 1$ to represent the ratio of outliers. 

We estimate the motion parameter $\theta$ and model points $X$ by maximizing the following log-likelihood,
\begin{equation}
    L = \sum_{i=1}^{M}\text{log}(\sum_{j=1}^{N+1}P(y_j)p(x_i(\theta)|y_j))
\end{equation}
\noindent here we restrict ourselves to the kinematic model $X = X(\theta)$ and leave the general case to supplemental materials. We use the EM~\cite{dempster1977em} algorithm to solve this optimization. The EM procedure is

\noindent \textbf{E step}: For each $x_i$, compute 
\begin{equation}
\label{equ:Estep}
\begin{aligned}
M^0_{x_i} &= \sum_{y_k} \mathcal{N}(x_i(\theta^{\text{old}});y_k, \Sigma_{xyz}) \\[-3pt]
M^1_{x_i} &= \sum_{y_k} \mathcal{N}(x_i(\theta^{\text{old}});y_k, \Sigma_{xyz}) y_k
\end{aligned}
\end{equation}
\noindent \textbf{M step}: minimize the following objective function 
\begin{equation}
\label{equ:Mstep}
\sum_{x_i} \frac{M^0_{x_i}}{M^0_{x_i} + c} (x_i(\theta) - \frac{M^1_{x_i}}{M^0_{x_i}})^{T} \Sigma_{xyz}^{-1} (x_i(\theta) - \frac{M^1_{x_i}}{M^0_{x_i}})
\end{equation}

\noindent where $M^0_{x_i}$ and $M^1_{x_i}$ are computed in the E step (\ref{equ:Estep}), $c = \frac{w}{1 - w} \frac{N}{M}$ is a constant, and $w$ is the parameter that represents the ratio of outliers. 

The EM procedure is conceptually related to ICP. The weight-averaged target point $({M^1_{x_i}}/{M^0_{x_i}})$ replaces the nearest neighbour in ICP, and each model point is weighted by $\frac{M^0_{X_i}}{M^0_{X_i} + c}$. Intuitively, the averaged target provides robustness to noise in observation, while the weight for each model point should reject outliers in the model. Please refer to supplemental materials for the complete derivation.

\subsection{Discussion and Comparison}
\label{subsec:comparison}

\begin{figure}[t]
\centering
\includegraphics[width=0.35\paperwidth]{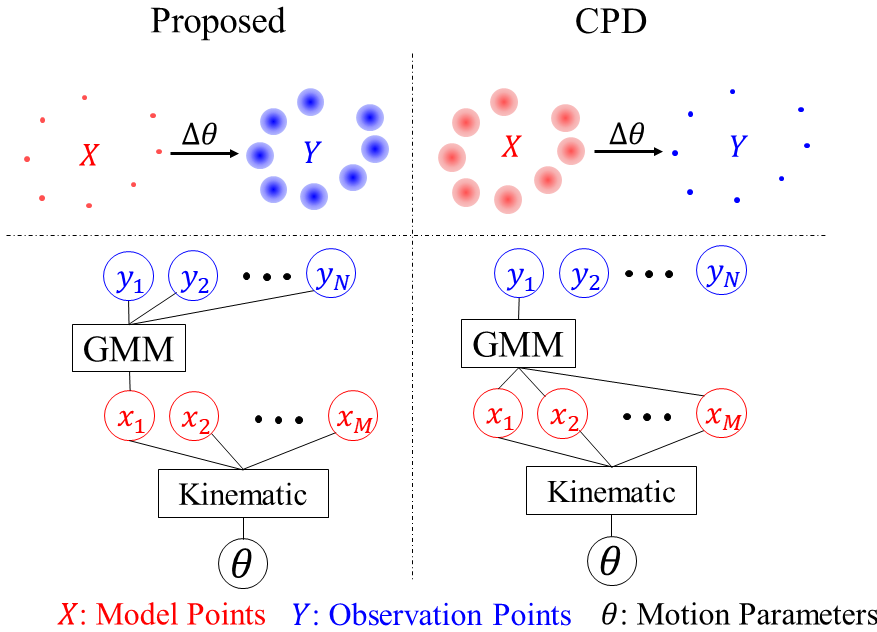}
\caption{\label{fig:formulation} An illustration of the proposed probabilistic model. Top: at a high level, the proposed formulation assumes the observation $Y$ introduces a probabilistic distribution, while CPD~\cite{cpd10} assumes the model $X$ introduces a distribution controlled by the motion parameter $\theta$. Bottom: factor graph representations of both our formulation and the formulation of CPD~\cite{cpd10}. }
\vspace{-0.7em}
\end{figure}

At a high level, the proposed formulation can be viewed as an ``inverse" of Coherent Point Drift (CPD)~\cite{cpd10} and many similar formulations~\cite{evangelidis2014generative, schulman2013tracking, condem2011}, as shown in Fig.~\ref{fig:formulation}. CPD~\cite{cpd10} assumes the observation points are independently distributed according to a GMM introduced by model points, while the proposed formulation directly assumes the observation points induce a GMM over the space. 
Empirically, both methods are very robust to noise and outliers and significantly outperform ICP.  

On the perspective of computation, the proposed method is much more simple and efficient than CPD~\cite{cpd10} and similar formulations~\cite{evangelidis2014generative, schulman2013tracking, condem2011}. The proposed method only requires sum over $Y$ (\ref{equ:Estep}), while CPD~\cite{cpd10} requires sum over both $Y$ and $X$. Moreover, if a spatial index is used to perform this sum, CPD~\cite{cpd10} must rebuild the index every EM iteration as the model points $X$ are updated. In our formulation, we only need to build the index once if the variance is fixed during EM iterations, which is sufficient for many applications~\cite{schulman2013tracking, wang2015deformation}.

Several existing works~\cite{treeguass2018, magnusson2009three} also build a GMM representation of the observation points. Compared with our method, they do not explicitly account for the outlier distribution and miss the weight $\frac{M^0_{X_i}}{M^0_{X_i} + c}$. Furthermore, these methods assume each model point is only correlated with one or several ``nearest" GMM centroids, while conceptually we assume each model point is correlated with all observation GMM centroids. 
Additionally, combined with the filter-based correspondence and twist parameterization in Sec.~\ref{sec:estep} and Sec.~\ref{sec:mstep}, our method tends to be much faster than these works, as demonstrated by our experiments.

\subsection{Several Extensions}
\label{subsec:extension}

The presented probabilistic formulation can be extended to incorporate many well-established GMM-based registration techniques.
Additionally, these extensions can be efficiently computed in a unified framework using the filter-based E step in Sec.~\ref{sec:estep} and the twist-based M step in Sec.~\ref{sec:mstep}.
We select the optimized variance proposed in~\cite{cpd10}, feature correspondence in~\cite{schulman2013tracking} and point-to-plane residual in~\cite{tricp2005} as three practically important examples, although many other methods can also be integrated in a very similar way.

\vspace{0.3em}
\noindent \textbf{Features}: Currently in the E step (\ref{equ:Estep}), only the 3d position is used to measure the similarity between the model and observation points. Similar to~\cite{schulman2013tracking}, the E step can be extended to incorporate features such as normal, SHOT~\cite{tombari2010shot}, learned features~\cite{schmidt2017self} or their concatenation. The E step for arbitrary feature is
\begin{equation}
\begin{aligned}
\label{equ:feature_estep}
M^0_{x_i} &= \sum_{y_k} \mathcal{N}(f_{x_i};f_{y_k}, \Sigma_{f}) \\[-3pt]
M^1_{x_i} &= \sum_{y_k} \mathcal{N}(f_{x_i};f_{y_k}, \Sigma_{f}) y_k
\end{aligned}
\end{equation}
\noindent where $f_{x_i}$ and $f_{y_k}$ are the feature value for point $x_i$ and $y_k$, $\Sigma_f$ is the diagonal covariance for the feature. 

\vspace{0.3em}
\noindent \textbf{Optimized Variance}: In our previous formulation, the variance of Gaussian kernel $\Sigma_{xyz}$ is used as a fixed parameter. Similar to CPD~\cite{cpd10}, if $\Sigma_{xyz} = \text{diag}(\sigma^2, \sigma^2, \sigma^2)$, the variance $\sigma$ can be interpreted as a decision variable and optimized analytically. Please refer to supplemental materials for the detailed formula and derivation.

\noindent \textbf{Point-to-Plane Distance}: The objective in our M step (\ref{equ:Mstep}) is similar to the point-to-point distance in ICP, which doesn't capture the planar structure. 
A simple solution is to compute a normal direction to characterize the local planar structure in the vicinity of the target $({M^1_{x_i}}/{M^0_{x_i}})$
\begin{equation}
\label{equ:pt2pl_estep}
N_{x_i} = (\sum_{y_k} \mathcal{N}(x_i;y_k, \Sigma_{xyz}) N_{y_k}) / M^0_{x_i}
\end{equation}
\noindent where $N_{y_k}$ is the normal of the observation point $y_k$. The objective in the M step is then a point-to-plane error
\begin{equation}
\label{equ:pt2pl_mstep}
\sum_{x_i} \frac{M^0_{x_i}}{M^0_{x_i} + c} \text{dot}(N_{x_i}, x_i(\theta) - \frac{M^1_{x_i}}{M^0_{x_i}})^{2}
\end{equation}

\section{E Step: Filter-based Correspondence}
\label{sec:estep}
\subsection{General Formulation}

In this section, we discuss the method to compute the E step (\ref{equ:Estep}) and several extensions (\ref{equ:feature_estep} and \ref{equ:pt2pl_estep}).
These specific E steps can be written into the following generalized form
\begin{equation}
\label{equ:gtransform}
G(f_{x_i}) = \sum_{y_k} e^{-\frac{1}{2}(f_{x_i} - f_{y_k})^{2}} v_{y_k}
\end{equation}
\noindent where $v_{y_k}$ generalizes the 3d position $y_k$ and the unit weight in (\ref{equ:Estep}, \ref{equ:feature_estep}) and the normal $N_{y_k}$ in (\ref{equ:pt2pl_estep}). The $G(f_{x_i})$ generalizes $M_{x_i}^{0}$ and $M_{x_i}^{1}$ in (\ref{equ:Estep}, \ref{equ:feature_estep}) and the normal $N_{x_i}$ in (\ref{equ:pt2pl_estep}).
The features $f_{x_i}$ and $f_{y_k}$ generalize 3d positions $x_i$ and $y_k$ in the Gaussian PDF $\mathcal{N}(x_i;y_k, \Sigma_{xyz})$. The features $f_{x_i}$ and $f_{y_k}$ are normalized to have identity covariance. We also omit the normalization constant $\text{det}(2\pi\Sigma_{xyz})^{-\frac{1}{2}}$ of the Guassian PDF $\mathcal{N}(x_i;y_k, \Sigma_{xyz})$ for notational simplicity. 

Equ.~(\ref{equ:gtransform}) is known as the general Gaussian Transform and the Improved Fast Gaussian Transform (IFGT)~\cite{yang2003improved} is proposed for it. However, IFGT~\cite{yang2003improved} uses a k-means tree internally and there would be too many k-means centroids for typical parameters in the registration. Practically, ~\cite{yang2003improved} is not much faster than brute-force evaluation for our task.

We instead propose to compute (\ref{equ:gtransform}) using Gaussian filtering algorithms~\cite{chen2007bilateralgrid, adams2010permu, adams2009gaussian}, which demonstrate promising accuracy and efficiency on image processing. The filtering operation that these algorithms accelerated is
\begin{equation}
\label{equ:gfilter}
G(f_{y_i}) = \sum_{y_k} e^{-\frac{1}{2}(f_{y_i} - f_{y_k})^{2}} v_{y_k}
\end{equation}
\noindent which is a subset of the general Gaussian transform: the feature $f_{y_i}$ used to retrieve the filtered value $G(f_{y_i})$ must be included in the input point set $Y$.

In our case, we would like to retrieve the value $G(f_{x_i})$ using feature $f_{x_i}$ from another point cloud $X$, which cannot be directly expressed in (\ref{equ:gfilter}). To resolve it, we propose the following augmented input:
\begin{equation}
\begin{aligned}
    F_{\text{filter-input}} &= [F_X, &F_Y] \\
    V_{\text{filter-input}} &= [0, &V_Y]
\end{aligned}
\end{equation}
\noindent where $F_X = [f_{x_1}, f_{x_2}, ..., f_{x_M}]$, $F_Y = [f_{y_1}, f_{y_2}, ..., f_{y_N}]$ and $V_Y = [v_{y_1}, v_{y_2}, ..., v_{y_N}]$. The new input feature $F_{\text{filter-input}}$ and value $V_{\text{filter-input}}$ are suitable for these filtering algorithms~\cite{chen2007bilateralgrid, adams2010permu, adams2009gaussian}, and the filtered output is
\begin{equation}
\begin{aligned}
G(f_{x_i}) &= \sum_{z_k \in F_{\text{filter-input}}} e^{-\frac{1}{2}(f_{x_i} - f_{z_k})^{2}} v_{z_k} \\
&= \sum_{y_k \in F_Y} e^{-\frac{1}{2}(f_{x_i} - f_{y_k})^{2}} v_{y_k}
\end{aligned}
\end{equation}

With this augmentation, we can apply these filtering algorithms~\cite{adams2010permu, adams2009gaussian, chen2007bilateralgrid} as a black box to our problem. However, by exploiting the structure of these methods, we can make them more efficient for our tasks. 
In the following text, the permutohedral lattice filter~\cite{adams2010permu} is discussed as an example, which is used in our experiments. 

\begin{figure}
\centering
\includegraphics[width=0.35\paperwidth]{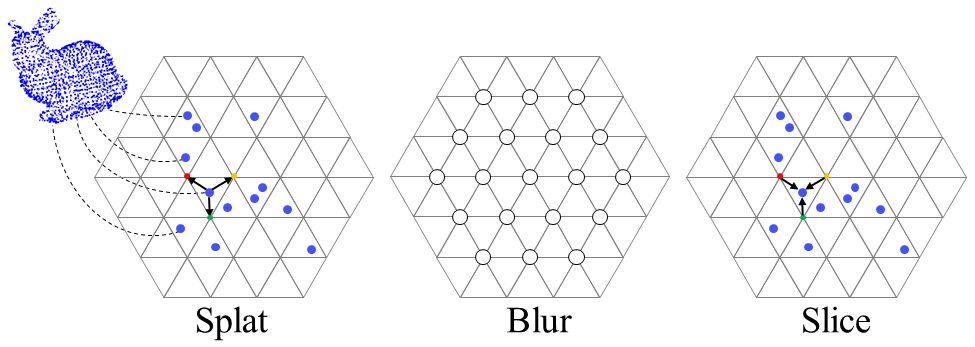}
\caption{\label{fig:lattice} An illustration of the permutohedral lattice filter~\cite{adams2010permu}. \textbf{Splat}: The input features are interpolated to permutohedral lattice using barycentric weights. \textbf{Blur}: lattice points exchange their values with nearby lattice points. \textbf{Slice}: The filtered
signal is interpolated back onto the input signal. }
\end{figure}

\subsection{Permutohedral Lattice Filter}
\label{subsec:permuto}

We briefly review the filtering process of ~\cite{adams2010permu}, an illustration is shown in Fig.~\ref{fig:lattice}. The $d$-dimension feature $f$ is first embedded in $(d+1)$-dimensional space, where the permutohedral lattice lives. In the embedded space, each input value $v$ \textbf{Splats} onto the vertices of its enclosing simplex with barycentric weights. Next, lattice points \textbf{Blur} their values with nearby lattice points. Finally, the space is \textbf{Sliced} at each input position using the same barycentric weights to interpolate output values. 

Although the permutohedral filter~\cite{adams2010permu} has demonstrated promising performance on a variety of tasks, it is still not optimal for our problem. In particular, the index building in~\cite{adams2010permu} can be inefficient when the variance $\Sigma_{xyz}$ is too small. Additionally, naively apply~\cite{adams2010permu} to the E step (\ref{equ:Estep}) requires rebuilding the index every EM iteration as the model point $X$ is updated. To resolve these problems, we propose a customization of the permutohedral filter~\cite{adams2010permu} that is more efficient while retaining sufficient accuracy for our task. The detailed method is presented in the supplemental material.

\section{M Step: Efficient Twist Parameterization}
\label{sec:mstep}

In this section, we present methods to solve the optimizations (\ref{equ:Mstep}, \ref{equ:pt2pl_mstep}) with the twist parameterization. We first discuss the twist in the general kinematic model, then specialize it to articulated and node-graph deformable objects.

We focus on the following general kinematic model,
\begin{equation}
\label{equ:kinematic}
x_i = T_i(\theta) x_{i\_\text{reference}}
\end{equation}

\noindent where $T_i(\theta) \in SE(3)$ is a rigid transformation, $x_{i\_\text{reference}}$ is the fixed reference point for the $x_i$. Note that $T_i(\theta)$ depends on $i$ and the kinematic model is not necessarily a global rigid transformation. 

Twist is a 6-vector that represents the locally linearized ``change" of $SE(3)$. Let the twist $\zeta_i = (w_i, t_i) = (\alpha_i, \beta_i, \gamma_i, a_i, b_i, c_i)$ be the local linearization of $T_i$, we have
\begin{equation}
T_i^{\text{new}} \approx 
\begin{bmatrix}
1 & -\gamma_i & \beta_i & a_i \\ \gamma_i & 1 & -\alpha_i & b_i \\ -\beta_i & \alpha_i & 1 & c_i \\ 0 & 0 & 0 & 1
\end{bmatrix}
T_i
\end{equation}
\noindent Thus, the Jacobian $\frac{\partial x_i}{\partial \zeta_i} = [\text{skew}(x_i), I_{3\times3}]$ is a $3\times6$ matrix, where $I_{3\times3}$ is identity matrix, and $\text{skew}(x_i)$ is a $3\times3$ matrix such that $\text{skew}(x_i) b = \text{cross}(x_i, b)$ for arbitrary $b \in {R}^3$.

The optimization (\ref{equ:Mstep}, \ref{equ:pt2pl_mstep}) are least squares problems, and we focus on the following generalized form of them
\begin{equation}
\label{equ:absM}
    \sum_{x_i} r_{x_i}^{T}r_{x_i}
\end{equation}
\noindent where $r_{x_i}$ is the concatenated least-squares residuals that depends on $x_i$. We use the Gauss-Newton (GN) algorithm to solve (\ref{equ:absM}). In each GN iteration we need to compute the following $A$ and $b$ matrices by
\begin{equation}
A = \sum_{x_i} (\frac{\partial {r_{x_i}}}{\partial \theta})^{T}\frac{\partial {r_{x_i}}}{\partial \theta},~~~ b = \sum_{x_i} (\frac{\partial {r_{x_i}}}{\partial \theta})^{T}(r_{x_i})
\end{equation}
\noindent and the update of the motion parameters is $\Delta \theta = -A^{-1} b$. Thus, the primary computational bottleneck is to assemble the matrices $A$ and $b$. In the following text, we only discuss the computation of the $A$ matrix, while the computation of the $b$ vector is similar and easier. The computation of the $A$ matrix can be written as 
\begin{equation}
A = \sum_{x_i} (\frac{\partial \zeta_i}{\partial \theta})^{T}((\frac{\partial {r_{x_i}}}{\partial \zeta_i})^{T}\frac{\partial {r_{x_i}}}{\partial \zeta_i}) (\frac{\partial \zeta_i}{\partial \theta})
\end{equation}
\noindent where $\frac{\partial \zeta_i}{\partial \theta}$ is the Jacobian that maps the change of motion parameter $\theta$ to the change of the rigid transformation $T_i$, while the change of $T_i$ is expressed as its twist. Note that the term $\frac{\partial {r_{x_i}}}{\partial \zeta_i} = \frac{\partial {r_{x_i}}}{\partial x_i} \frac{\partial x_i}{\partial \zeta_i}$ is very easy to compute, as both $\frac{\partial {r_{x_i}}}{\partial x_i}$ and $\frac{\partial x_i}{\partial \zeta_i}$ are only dependent on $x_i$.

If the kinematic model is a global rigid transformation, we have $\frac{\partial \zeta_i}{\partial \theta} = I_{6\times6}$ and $A = \sum_{x_i} ((\frac{\partial {r_{x_i}}}{\partial \zeta_i})^{T}\frac{\partial {r_{x_i}}}{\partial \zeta_i})$. In the following subsections, we proceed to the articulated and node-graph deformable kinematic models.

\subsection{Articulated Model}

Articulated objects consist of rigid bodies connected through joints in a kinematic tree. A broad set of real-world objects, including human bodies, hands and robots are articulated objects. If the kinematic model (\ref{equ:kinematic}) is an articulated model, the motion parameter $\theta \in R^{N_{\text{joint}}}$ would be the joint angles, where $N_{\text{joint}}$ is the number of joints. The $T_i(\theta)$ is the rigid transform of the rigid body that the point $x_i$ is attached to. The computation of the $A$ matrix can be factored as
\begin{equation}
\label{equ:articulated}
A = \sum_{\text{body}_j} (\frac{\partial \zeta_j}{\partial \theta})^{T}(\sum_{x_i \text{ in body}_j} (\frac{\partial {r_{x_i}}}{\partial \zeta_i})^{T}\frac{\partial {r_{x_i}}}{\partial \zeta_i}) (\frac{\partial \zeta_j}{\partial \theta})
\end{equation}
\noindent where $\zeta_j$ is the twist of rigid body $j$, and we have exploited $\frac{\partial \zeta_i}{\partial \zeta_j} = I_{6\times6}$ if point $i$ is on rigid body $j$. Importantly, $\frac{\partial \zeta_j}{\partial \theta}$ is known as the spatial velocity Jacobian and is provided by many off-the-shelf rigid body simulators~\cite{drake, todorov2012mujoco, lee2018dart}. The algorithm that uses (\ref{equ:articulated}) is shown in Algorithm~\ref{alg:articulated}. 

The lines 1-4 of Algorithm~\ref{alg:articulated} dominates the overall performance and the complexity is O($6^2M$), where $M$ is the number of model points and usually $M \gg N_{\text{joint}}$. Thus, the complexity of this algorithm is almost independent of $N_{\text{joint}}$. As a comparison, previous articulated registration methods~\cite{lee2018dart, tkach2016sphere} need O($N_{\text{joint}}^{2}M$) time to assemble the $A$ matrix, and $N_{\text{joint}}$ is usually much larger than 6. Furthermore, lines 1-4 of Algorithm~\ref{alg:articulated} is very simple to implement and can be easily GPU parallelized. Combined with an off-the-shelf simulator, the overall pipeline can achieve promising efficiency. On the contrary, previous methods~\cite{lee2018dart, tkach2016sphere} typically need a customized kinematic tree implementation for real-time performance, 
while requires substantial software engineering effort to realize.

\algnewcommand{\LineComment}[1]{\State \(\triangleright\) #1}

\begin{algorithm}[t]
\caption{The $A$ matrix for articulated kinematic}\label{alg:articulated}
\begin{algorithmic}[1]
\For{all $\text{body}_j$}\Comment{can be parallelized}
\State ${J^TJ}_{\text{twist}\_j} = 0_{6\times6}$
\For{all point $x_i$ in $\text{body}_j$}\Comment{can be parallelized}
\State ${J^TJ}_{\text{twist}\_j} \mathrel{{+}{=}} (\frac{\partial {r_{x_i}}}{\partial \zeta_i})^{T}\frac{\partial {r_{x_i}}}{\partial \zeta_i}$
\EndFor
\EndFor

\State $A = 0_{N_{\text{joint}}\times N_{\text{joint}}}$
\For{all $\text{body}_j$}
\LineComment{The spatial velocity Jacobian can be computed}
\LineComment{using off-the-shelf simulators such as \cite{todorov2012mujoco, drake}}
\State $J_{\text{spatial}\_j}=$ spatial velocity Jacobian of $\text{body}_j$
\State $A \mathrel{{+}{=}} J_{\text{spatial}\_j}^{T} ({J^TJ}_{\text{twist}\_j}) J_{\text{spatial}\_j}$
\EndFor

\end{algorithmic}
\end{algorithm}

\subsection{Node-Graph Deformable Model}

To capture the motion of objects such as rope or cloth, we need a kinematic model which allows large deformation while preventing unrealistic collapsing or distortion. 
In this paper, we follow~\cite{kavan2006dual} to represent the general deformable kinematic model as a node graph. Intuitively, the node graph defines a motion field in the 3D space and the reference vertex in Equ.~(\ref{equ:kinematic}) is deformed according to the motion field. More specifically, the node graph is defined as a set $\{ [ p_{j} \in R^{3}, T_{j} \in SE(3) ]  \} $, where $j$ is the node index, $p_j$ is the position of the $j$th node, and $T_j$ is the $SE(3)$ motion defined on the $j$th node. 
The kinematic equation (\ref{equ:kinematic}) can be written as 

\vspace{-0.9em}
\begin{equation}
\label{equ:warping}
T_i(\theta) = \text{normalized}(\Sigma_{k \in N_i(x_{i\_\text{reference}})} w_{ki} T_k)
\end{equation}

\noindent where $N_i(x_{i\_\text{reference}})$ is the nearest neighbor nodes of model point $x_{i\_\text{reference}}$, and $w_{ki}$ is the fixed skinning weight. The interpolation of the rigid transformation $T_k$ is performed using the DualQuaternion~\cite{kavan2008geometric} representation of the $SE(3)$. 

The $A$ matrices for this kinematic model can be constructed using an algorithm very similar to Algorithm~\ref{alg:articulated}. The detailed method is provided in supplemental materials. 

\section{Results}

We conduct a variety of experiments to test the robustness, accuracy and efficiency of the proposed method. Our hardware platform is an Intel i7-3960X CPU except for Sec.~\ref{subsec:deformable-result}, where the proposed method is implemented with CUDA on a Nivida Titan Xp GPU. The video demo and the source code are available on our  \href{https://sites.google.com/view/filterreg/home}{project page}.

\subsection{Robustness Test on Synthetic Data}


We follow CPD~\cite{cpd10} to setup an experiment on synthetic data. We use a subsampled Stanford bunny with 3500 points. 
The initial rotation discrepancy is 50 degrees with a random axis. The proposed method is compared with two baselines: CPD~\cite{cpd10}, a representative GMM-based algorithm; TrICP~\cite{tricp2005}, a widely used robust ICP variant. Parameters for all methods are well tuned and provided in supplemental materials.
We use the following metric to measure the pose estimation error
\begin{equation}
\label{equ:alignerror}
    \text{error}(T) = \frac{1}{M} \Sigma_{i=1}^{M} |(T - T_{\text{gt}}) x_{i\_\text{reference}}|_2
\end{equation}
\noindent where $T_{\text{gt}}$ is the known ground truth pose, $x_{i\_\text{reference}}$ defined in (\ref{equ:kinematic}) is the reference position. We terminate the algorithm when the twist (change of transformation) is less than a threshold. In this way, the final alignment error (\ref{equ:alignerror}) is about 1 [mm] for all methods. All of the statistical results 
are the averaged value of 30 independent runs.

\begin{figure}[t]
\centering
\includegraphics[width=0.3\paperwidth]{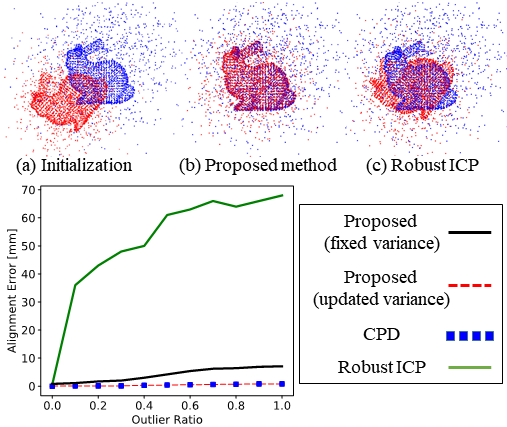}
\caption{\label{fig:bunnyoutlier} A comparison of the robustness of various algorithms with respect to outliers. Top: (a) shows an example initialization with 0.2 outlier ratio; (b) and (c) are the final alignment by the proposed method and TrICP~\cite{tricp2005} initialized from (a), respectively. Bottom: the alignment error (\ref{equ:alignerror}) of each algorithm for different numbers of outliers. }
\end{figure}

Fig.~\ref{fig:bunnyoutlier} shows the robustness of different algorithms with respect to outliers in the point sets. We add different number of points randomly to both the model and observation clouds. An example of such point sets with initial alignment is shown in Fig.~\ref{fig:bunnyoutlier} (a), the converged alignment by the proposed method and TrICP~\cite{tricp2005} are shown in Fig.~\ref{fig:bunnyoutlier} (b) and Fig.~\ref{fig:bunnyoutlier} (c), respectively. The proposed method and CPD~\cite{cpd10} significantly outperform the robust ICP. 

\begin{figure}[t]
\centering
\includegraphics[width=0.3\paperwidth]{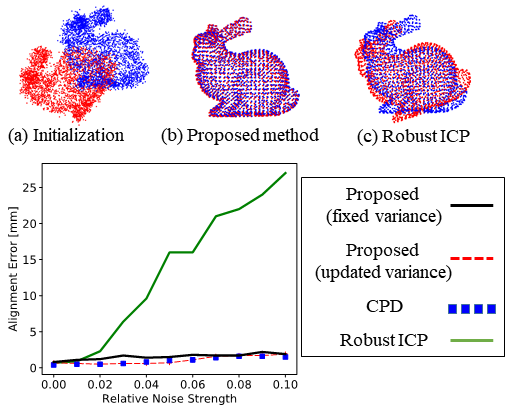}
\caption{\label{fig:bunnynoise} A comparison of the robustness of various algorithms with respect to noise. Top: (a) shows an example initialization with 0.03 relative noise; (b) and (c) are the final alignment by the proposed method and TrICP~\cite{tricp2005} initialized from (a). Note that we use clean point clouds for better visualization. Bottom: the alignment error (\ref{equ:alignerror}) of each algorithm for different levels of noise. }
\end{figure}

Fig.~\ref{fig:bunnynoise} shows the robustness of different algorithms with respect to noise in the point sets. We corrupt each point in both model and observation clouds with a Gaussian noise. The unit of the noise is the diameter of the Bunny. An example of such point sets with initial alignment are shown in Fig.~\ref{fig:bunnynoise}~(a). Fig.~\ref{fig:bunnynoise} (b) and (c) are the final alignment by the proposed method and TrICP~\cite{tricp2005} initialized from (a). Note that we use clean point clouds for better visualization. Our method and CPD~\cite{cpd10} are more accurate than the robust ICP.

Table.~\ref{table:bunnyspeed} summarizes the computational performance of each method. The running time is measured on clean point cloud. 
Our method is about 7 times faster than TrICP~\cite{tricp2005} and two orders of magnitude faster than CPD~\cite{cpd10}. The proposed method with fixed $\sigma$ is faster per iteration, but need more iterations to converge. Overall the proposed method is as robust as the state-of-the-art statistical registration algorithm CPD~\cite{cpd10}, and runs substantially faster than the modern ICP implementation. 

\begin{table}[]
\centering
\label{table:bunnyspeed}
\begin{tabular}{|c|c|c|c|}
\hline
\multicolumn{1}{|l|}{}                                     & \begin{tabular}[c]{@{}c@{}}time{[}ms{]}\\ per iteration\end{tabular} & \#iterations & \begin{tabular}[c]{@{}c@{}}overall \\ time{[}ms{]}\end{tabular} \\ \hline
\begin{tabular}[c]{@{}c@{}}Proposed \\ fixed $\sigma$ \end{tabular}  & \textbf{0.96}                                                                 & 40.4         & \textbf{38.4}                                                            \\ \hline
\begin{tabular}[c]{@{}c@{}}Proposed\\ updated $\sigma$ \end{tabular} & \textbf{1.16}                                                                 & \textbf{27.6}         & \textbf{32.1}                                                            \\ \hline
CPD                                                        & 228                                                                  & \textbf{26.8}         & 6110                                                            \\ \hline
Robust ICP                                                 & 3.10                                                                 & 70.2         & 217.6                                                           \\ \hline
\end{tabular}
\caption{The performance of different algorithms for the registration of the Stanford Bunny. }
\end{table}

\subsection{Rigid Registration on Real-World Data}
\label{subsec:rigid_real}

\begin{figure}
\centering
\includegraphics[width=0.4\paperwidth]{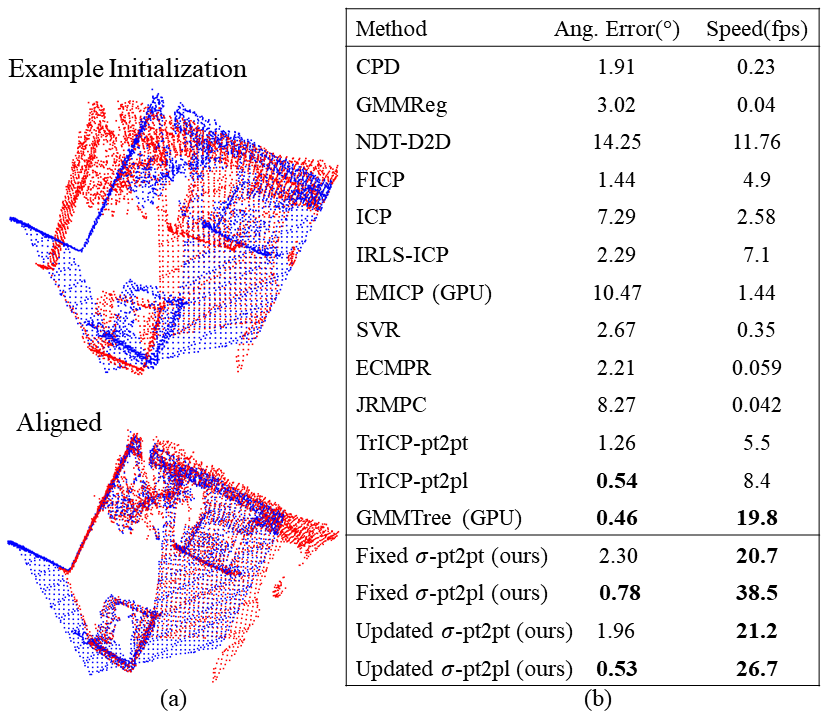}
\caption{\label{fig:lounge} Rigid registration on the Stanford Lounge dataset \cite{zhou2013dense}. The results for most baselines are from~\cite{treeguass2018}. (a) shows an example registration by the proposed method. (b) shows the accuracy and performance of various algorithms. 
}
\end{figure}

We follow \cite{treeguass2018} to setup this experiment: the algorithm is used to compute the frame-to-frame rigid transformation on the Stanford Lounge dataset \cite{zhou2013dense}. We register every 5th frame for the first 400 frames, each downsampled to about 5000 points. The average Euler angle deviation from the ground truth is used as the estimation error. 

Fig.~\ref{fig:lounge} (a) shows an example registration by the proposed method. Fig.~\ref{fig:lounge} (b) shows the accuracy and speed of various algorithms. The results of baseline methods are from \cite{treeguass2018}\footnote[1]{Our CPU (i7-3960X) is slightly inferior to \cite{treeguass2018} (i7-5920K), and we observe similar accuracy and slightly worse speed using CPD~\cite{cpd10} and TrICP~\cite{tricp2005}. Thus, we think our speed result are comparable to~\cite{treeguass2018} despite hardware difference.} except for CPD~\cite{cpd10}. For CPD~\cite{cpd10} we use $\sigma_{\text{init}} = 20\text{ [cm]}$ instead of the data-based initialization of~\cite{cpd10}, with which we observed improved performance. As the point-to-point error doesn't capture the planar structure, the point-to-point version of the proposed method as well as many other point-to-point algorithms~\cite{cpd10, magnusson2009three, icp1992, tricp2005} are less accurate on this dataset. The proposed method with point-to-plane error achieves state-of-the-art accuracy. On the perspective of computation, the proposed method significantly outperforms all the baselines, including GMMTree~\cite{treeguass2018} and EMICP~\cite{granger2002multi} that rely on a high-end Titan X GPU.  

\subsection{Global Pose Estimation using Learned Features}

\begin{figure}
\centering
\includegraphics[width=0.38\paperwidth]{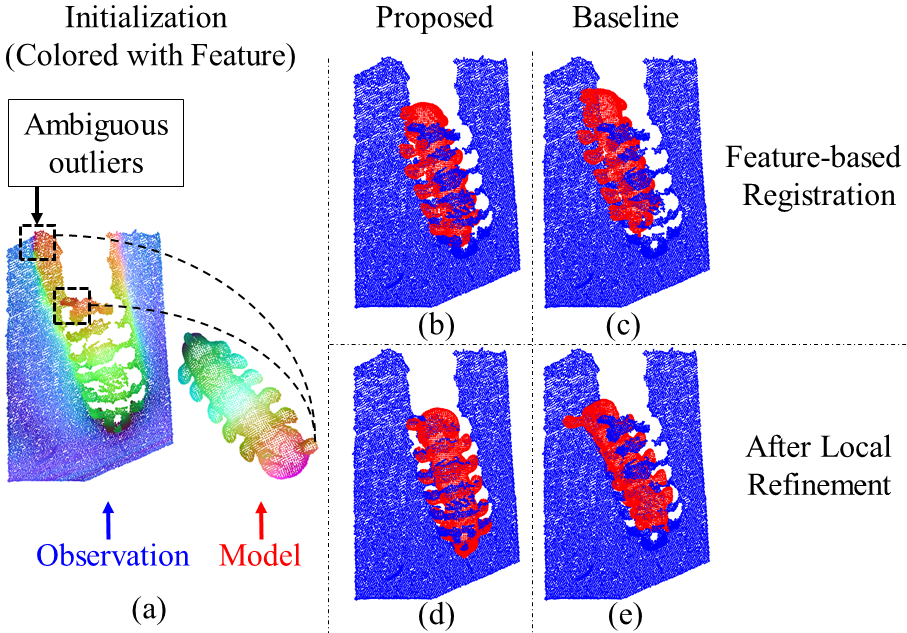}
\caption{\label{fig:nosegment} A feature-based global registration under ambiguous outliers and strong occlusion. (a) shows the initialization of the registration colored by the feature~\cite{don2018}. (b) and (c) 
are the feature-based registration by our method and the baseline.
(d) and (e) show the final alignment using 3d local refinement initialized from (b) and (c). The proposed method converges to the correct pose while the baseline method is trapped to bad alignment. }
\vspace{-1em}
\end{figure}

We demonstrate global pose estimation using motion-invariant features. 
The task is to align a pre-built geometric model to observation clouds from RGBD images, where both the model and observation clouds are colored by the learned feature~\cite{don2018}. We use the proposed method with feature correspondence in Sec.~\ref{subsec:extension} and fixed $\sigma = 0.05$ as the feature has unit norm. The proposed method is compared with a modified TrICP~\cite{tricp2005}: the nearest neighbour is searched in feature space (instead of 3d-space). After feature-based registration, we apply 3d-space local refinement to get the final alignment. 


Fig.~\ref{fig:nosegment} shows an example registration. Note that we treat the background as outliers. 
As shown in Fig.~\ref{fig:nosegment} (a), the observation (RGBD cloud) is under severe occlusion and contains very ambiguous outliers. 
Fig.~\ref{fig:nosegment} (b) and (c) show feature-based registration by our method and the ``feature" TrICP. The proposed method is more robust to the outliers and occlusion. Fig.~\ref{fig:nosegment} (d) and (e) show the final alignment using local refinement initialized from (b) and (c). The proposed method converges to correct pose while the baseline is trapped to bad alignment. Table.~\ref{table:feature} summaries the success rate of both methods on 30 RGBD images with different view points and lighting conditions. Our method has a higher success rate and is more efficient than the baseline.

\begin{table}
\centering
\begin{tabular}{|c|c|c|}
\hline
            & success rate & time {[}ms{]} \\ \hline
Proposed    & 29/30        & 13            \\ \hline
Feature ICP & 25/30        & 34            \\ \hline
\end{tabular}
\caption{\label{table:feature}The success rate and speed on the feature-based global registration. }
\vspace{-1em}
\end{table}

\subsection{Articulated Tracking}
The proposed method with articulated kinematic model is used to track a robot manipulating a box. The robot and box model has 20 DOFs (12 for the floating bases of the box and the robot, 8 for robot joints). We use drake~\cite{drake} for the kinematic computation in (\ref{equ:articulated}). We use fixed $\sigma = 1$~[cm] and set the maximum EM iterations to be 15. Our template has 7500 points and the depth cloud has about 10000 points.

\begin{figure}[t]
\centering
\includegraphics[width=0.37\paperwidth]{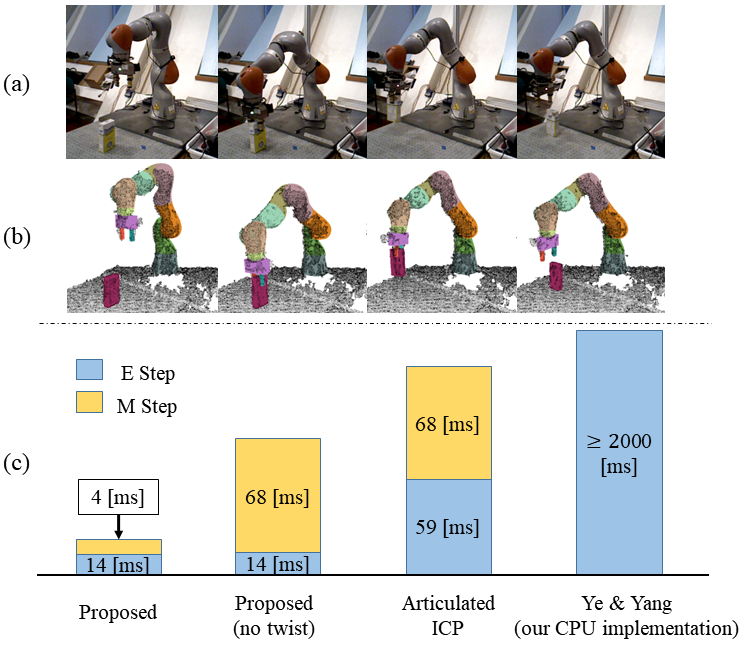}
\caption{\label{fig:articulated} The proposed method is applied to track a robot manipulating a box. (a): the snapshots of the tracked manipulation scenario. (b) the observation point clouds (black) and the live geometric model (colored according to rigid bodies). (c): the per-frame performance of various algorithms on this dataset. Ye \& Yang stands for our CPU implementation of ~\cite{ye2014real}. }
\end{figure}

Fig.~\ref{fig:articulated} (a) shows the snapshots of the tracked manipulation scenario. Fig.~\ref{fig:articulated} (b) shows the live geometric model and the observation clouds. Points from observation are in black, while the geometric model is colored according to rigid bodies. Note that points from the table are treated as outliers. Fig.~\ref{fig:articulated} (c) summaries the averaged per-frame performance of various algorithms. The proposed twist parameterization is an order of magnitude faster than direct parameterization. Combining the filter-based correspondence and twist parameterization leads to a real-time tracking algorithm and substantial performance improvement over articulated ICP and ~\cite{ye2014real}. 

\subsection{Application to Dynamic Reconstruction}
\label{subsec:deformable-result}

\begin{figure}[h]
\centering
\includegraphics[width=0.33\paperwidth]{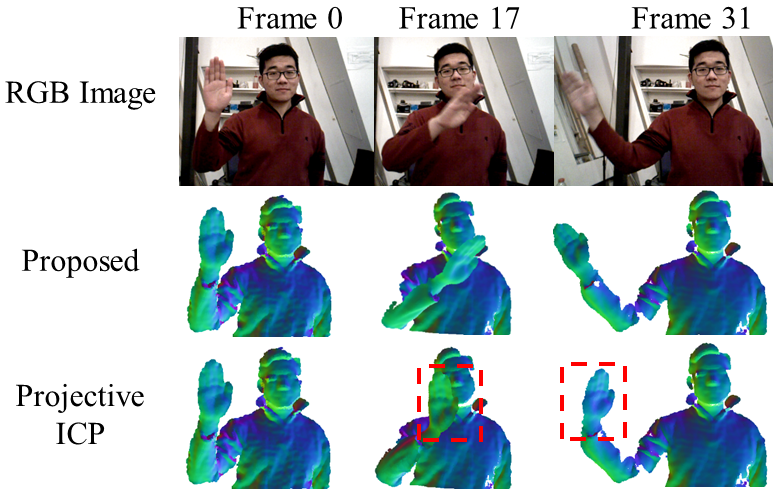}
\caption{\label{fig:deformable} The proposed method with node-graph deformable kinematic is implemented on GPU and used as the internal non-rigid tracker of DynamicFusion~\cite{newcombe2015dynamicfusion}. For a relative fast motion, the proposed method tracks it correctly while the projective ICP used by DynamicFusion~\cite{newcombe2015dynamicfusion} fails to track the right hand of the actor.  }
\end{figure}

The proposed method with node-graph deformable kinematic is implemented on GPU and used as the internal non-rigid tracker of DynamicFusion~\cite{newcombe2015dynamicfusion} (our implementation). The proposed method is compared with the projective ICP, the original non-rigid tracker of~\cite{newcombe2015dynamicfusion}. We use fixed $\sigma = 2$ [cm]. Fig.~\ref{fig:deformable} shows both methods operate on a RGBD sequence with relative fast motions. The proposed method tracks it correctly, while the projective ICP fails to track the right hand of the actor. The proposed method is more robust to fast and tangential motion than the projective ICP.

To test the efficiency of the proposed twist parameterization on node-graph deformable objects, we compare it with Opt~\cite{devito2017opt}, a highly optimized GPU least squares solver using direct parameterization. The per-frame computational performance of various algorithms is summarized in Table.~\ref{table:deformable-speed}. The GPU parallelization of our filter-based E step achieves 8 times speedup over the CPU version, and the proposed twist parameterization is about 20 times faster than ~\cite{devito2017opt}.

\begin{table}[h]
\centering
\begin{tabular}{|c|c|c|c|}
\hline
                & \begin{tabular}[c]{@{}c@{}}Proposed\\ (GPU)\end{tabular} & \begin{tabular}[c]{@{}c@{}}Proposed \\ (CPU)\end{tabular} & \begin{tabular}[c]{@{}c@{}}Proposed \\ (Opt~\cite{devito2017opt}) \end{tabular} \\ \hline
E step {[}ms{]} & 7.8                                                      & 62                                                        & 7.8                                                            \\ \hline
M step {[}ms{]} & 21.6                                                     & Not implemented                                                        & 382                                                            \\ \hline
\end{tabular}
\caption{\label{table:deformable-speed} The per-frame performance of various algorithms for deformable tracking on the sequence in Fig.~\ref{fig:deformable}.  }
\vspace{-0.8em}
\end{table}

\vspace{-0.4em}
\section{Conclusion}
\vspace{-0.4em}

To conclude, we present a probabilistic registration method that achieves state-of-the-art robustness, accuracy and efficiency. We show that the correspondence search can be formulated as a filtering problem, and employ advances in efficient Gaussian filtering methods to solve it. In addition, we present a simple and efficient twist parameterization that generalizes our method to articulated and deformable objects. Extensive empirical evaluation demonstrates the effectiveness our method.

\noindent {\footnotesize \textbf{Acknowledgments} This work was supported by NSF Award IIS-1427050 and Amazon Research Award. The views expressed in this paper are those of the authors themselves and are not endorsed by the funding agencies.}

{\small
\bibliographystyle{ieee}
\bibliography{egbib.bib}
}

\end{document}